\documentclass[9pt, conference]{IEEEtran}

\usepackage{cite}
\usepackage{amsmath,amssymb,amsfonts}
\usepackage{comment}
\usepackage{hyperref}
\usepackage{textcomp}
\usepackage{xcolor}
\usepackage{algorithm2e}
\usepackage{enumitem}
\usepackage{multirow}
\def\BibTeX{{\rm B\kern-.05em{\sc i\kern-.025em b}\kern-.08em
    T\kern-.1667em\lower.7ex\hbox{E}\kern-.125emX}}

\ifCLASSINFOpdf
   \usepackage[pdftex]{graphicx}
    \usepackage{caption}
    \usepackage{subcaption}
 \graphicspath{{./Figures/}}
  \DeclareGraphicsExtensions{.pdf,.jpg,.png}
\else
\fi  
\newcommand{\etal}{\textit{et al.}}

\makeatletter
\DeclareRobustCommand\onedot{\futurelet\@let@token\@onedot}
\def\@onedot{\ifx\@let@token.\else.\null\fi\xspace}

\def\eg{\emph{e.g}\onedot}

\def\etal{\emph{et al}\onedot}
\makeatother

\begin{document}

\title{HDL-GPT: High-Quality HDL is All You Need}

\author{
\centerline{Bhuvnesh Kumar, Saurav~Nanda, Ganapathy Parthasarathy}\\
\centerline{Pawan Patil, Austin Tsai and Parivesh Choudhary}\\
\IEEEauthorblockN{
\textit{Synopsys Inc., Sunnyvale, CA} \\
(bhuvnesh, sauravn, gpartha, pawanp, chengyun, parivesh)$@$synopsys.com
}}
\maketitle

\begin{abstract}
This paper presents Hardware Description Language Generative Pre-trained Transformers (HDL-GPT), a novel approach that leverages the vast repository of open-source High Definition Language (HDL) codes to train superior quality large code models. The core premise of this paper is the hypothesis that high-quality HDL is all you need to create models with exceptional performance and broad zero-shot generalization abilities. The paper elucidates the methods employed for the curation and augmentation of large corpora from open-source HDL code, transforming highly variable quality data into high-quality data through careful prompting and context maintenance. We demonstrate that the careful selection, filtering, and augmentation of data across HDLs can yield powerful models that surpass current state-of-the-art models. We also explore the impact of different fine-tuning methods on the quality of results. We describe experimental results across a range of fine-tuned SOTA LLMs, substantiating our claims. We demonstrate improvements of $50\%$ to $200\%$ over SOTA HDL models on current benchmarks in tasks ranging from HDL circuit explanations, code generation, formal and simulation testbench creation, triaging bugs, and fixing them. HDL-GPT opens new avenues for the development of advanced model training techniques for circuit design tasks.
\end{abstract}

\section{Introduction}
The relentless pace of Moore's law~\cite{mack2015multiple} has ushered in a surge of technological advancements in the realm of integrated circuit (IC) design. EDA algorithms and tools have led this revolution, facilitating the creation of feature-rich System-on-Chip (SoC) designs with billions of transistors. However, both Moore’s law and increasing complex functionality requirements also necessitate constant advancements in chip design productivity.
Recently, Large Language Models (LLMs) have emerged as a promising tool to augment EDA processes. Researchers have focused on automating language-related chip design tasks, which often involve time-consuming interfacing with natural or programming languages. LLMs, both commercial (\eg OpenAI~\cite{openai2024gpt4technicalreport}) and open source (\eg Mixtral\cite{jiang2024mixtralexperts}, LLaMA2\cite{touvron2023llama2openfoundation}), have opened new avenues for automating these tasks. These LLMs can generate code, conduct analysis, and respond to engineering queries via a natural language interface.

However, there are significant challenges in deploying LLMs in Hardware Description Language (HDL) code generation. For example, Tsai \etal~\cite{tsai2024rtlfixerautomaticallyfixingrtl} found that up to 55\% of errors generated by LLMs in Verilog code are syntax errors. The complexity of HDL design extends beyond generating syntactically correct code and achieving precision often necessitates multiple iterations. The challenge lies in meeting performance, power, and area (PPA) requirements for the overall IC design, which requires eliminating syntactic, semantic, or tool-specific errors. This flow dependency of errors presents a significant obstacle to automating chip design tasks.

A crucial part of solutions to these challenges is the quality and diversity of training data for LLMs to understand both syntax and semantics and enable robust zero-shot learning. Additionally, exposure to diverse types of errors in training data helps LLMs learn to detect and rectify such errors in the generated code.

\begin{figure*}[t!hbp]
	\centering
	\includegraphics[width=.8\textwidth]{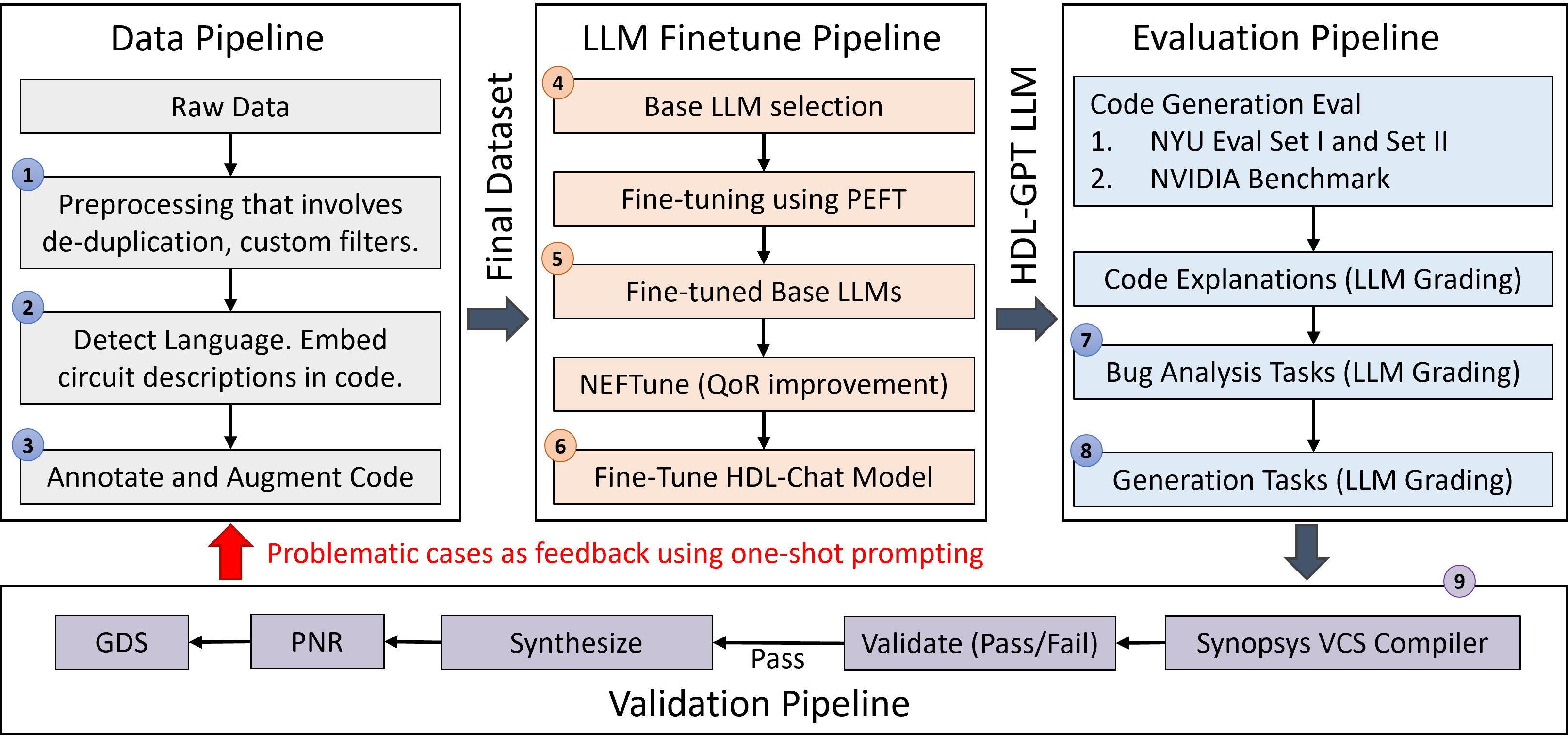}
    \vspace{-0.2cm}
\	\caption{Overall workflow for Data, Fine-tuning, Evaluation, and Verification/Feedback Pipeline for HDL-GPT}
	\label{fig_2}
 \vspace{-0.2cm}
\end{figure*}
There is significant prior art describing why high-quality training data is integral to the LLM training and fine-tuning. These include the following categories of motivating reasons:
\begin{enumerate}
\item \textbf{Quality of Generated Code:} LLM performance is directly tied to the quality of the training data. High-quality data aids in producing more accurate, efficient, and robust code. This also exposes LLMs to real-world coding scenarios and challenges, enabling practical solutions with LLMs. For example, Gunasekar et. al \cite{gunasekar2023textbooksneed} demonstrated the influence of high-quality data on enhancing a language model's efficiency in code-generation tasks.
\item \textbf{Understanding Context:} Training data facilitates comprehension of context in code use, crucial for generating semantically appropriate code.
\item \textbf{Syntax and Semantics:} Training data assists LLMs in understanding both syntax and semantics of the programming language, enabling them to learn from examples~\cite{thakur2024verigen, liu2024chipnemodomainadaptedllmschip}.
\item \textbf{Zero-shot Learning:} High-quality training data enables effective generalization in zero-shot learning scenarios, helping LLMs generalize to code for untrained tasks~\cite{li2023starcodersourceyou}.
\item \textbf{Error Handling:} Exposure to diverse errors through training data equips LLMs with the ability to manage and rectify errors in the generated code~\cite{tsai2024rtlfixerautomaticallyfixingrtl}.
\item \textbf{Dataset Size:} The size of the dataset also plays a crucial role in acquiring high-quality HDL data~\cite{hoffmann2022trainingcomputeoptimallargelanguage}. However, manually gathering a sufficiently large training corpus is a daunting task, both in terms of cost and effort. Therefore, the process needs to be automated for efficient data acquisition.
\end{enumerate}

This paper introduces a systematic automated approach to generating high-quality data for HDL generation and analysis tasks to address these challenges and harness the potential of LLMs in HDL-based IC design. We explain how we curate and augment large corpora from open-source HDL code, thereby transforming highly variable quality data into high-quality data through careful prompting and context maintenance. We use augmented data to train high-quality large code models with exceptional performance and broad zero-shot generalization abilities. We also explore the impact of different fine-tuning methods on the quality of results and compare the results with the current state of the art in the field.

\section{Automated HDL Data Augmentation}
Our approach to data augmentation involves a systematic process known as \textit{chain of thought prompting}~\cite{wei2022chain}. The overall process is shown in Figure~\ref{fig_2} as a series of task-specific pipelines with feedback. The figure shows a conceptual breakdown of the process into a Data Pipeline, an LLM Finetune Pipeline, an Evaluation Pipeline, and a feedback enabling Validation Pipeline. 

The Data Pipeline begins with data curation where we collect HDL code data from GitHub. We initially filter code repositories to remove code with non-permissible licenses. We also remove any code that matches the code or expected code in the evaluation data-sets to prevent data leakage during quality evaluations. The filtered data forms our initial raw data corpus. We go through well-known techniques such as de-duplication, and custom filters to remove objectionable material from the raw data corpus. Once we have a curated data corpus, we can address the ten major functional steps in the augmentation process shown in Table~\ref{tab:cot} as S1 through S10. 

\begin{table}[h!]  
\centering  
\caption{Data Augmentation with COT~\cite{wei2022chain} prompting}  
\begin{tabular}{|p{0.07\columnwidth}|p{0.85\columnwidth}|}  
\hline  
\textbf{Step} & \textbf{Task Description} \\  
\hline  
S1 & Identify the language and characteristics of the circuit. \\  
\hline  
S2 & Explain the functionality of the circuit. If it is combinational, generate a Karnaugh map or logic table. \\  
\hline  
S3 & Add clear descriptions and comments to the existing code. \\  
\hline  
S4 & Evaluate and rate the quality of the documented code. \\  
\hline  
S5 & Optimize the code considering potential improvements. \\  
\hline  
S6 & Generate a testbench for the optimized code. \\  
\hline  
S7 & Create SystemVerilog assertions for the optimized code. \\  
\hline  
S8 & Identify possible errors in the optimized code. \\  
\hline  
S9 & Generate a version of the circuit with a specific error. \\  
\hline  
S10 & Create a testbench to detect a specific error in the circuit. \\  
\hline  
S11 & Validate and score results of each task in the circuit. \\  
\hline  
\end{tabular}  
\label{tab:cot}
\end{table}  

The process begins with the identification of the circuit's language and characteristics, followed by a detailed explanation of its functionality. Existing code is then annotated with filename, clear descriptions and comments (step 2 in Figure~\ref{fig_2} mapping to S1-S4 in Table~\ref{tab:cot}), after which an evaluation of the code's quality is performed. The quality assessment is used to guide optimizations to the code. A testbench for the optimized code is then generated, followed by the creation of SystemVerilog assertions. We next identify possible errors in the optimized code. Subsequently, a version of the circuit with a specific error is generated for the error detection task and validation thereof. The final step involves creating a testbench to detect this specific error in the circuit (Step 3 in Figure~\ref{fig_2} maps to S5-S10 in Table~\ref{tab:cot}). Each step is performed using a complex prompt that is tuned to generate the best possible answers from SOTA LLMs such as Mixtral 8x7B~\cite{jiang2024mixtralexperts}. 

The final validation stage compiles the code samples using the Synopsys Verilog Compiler Simulator (VCS)~\footnote{https://www.synopsys.com/verification/simulation/vcs.html}. This stage runs each sample through the full EDA flow to GDS generation wherever applicable, categorizing them as accepted or rejected based on their compilation success and PPA results. The quality of generated data from each step is evaluated for syntactic and performance problems using Synopsys EDA tools and a student-teacher grading model to measure qualitative tasks such as comments and functionality. The results of the evaluations are used to score the generated data for fine-tuning and as feedback to the generation process to improve subsequent generation using one-shot prompting. Steps 7-9 in Figure~\ref{fig_2} maps to S11 in Table~\ref{tab:cot} for validation and scoring. The final dataset, after all augmentations, contains 1.31 billion tokens and is used to fine-tune all candidate models.

The LLM finetune pipeline selects and fine-tunes various Language Learning Models using the augmented data-set, including StarCoder 16B \cite{li2023starcodersourceyou} variants, CodeLlama 7B \cite{roziere2024codellamaopenfoundation}, through PEFT \cite{hu2021loralowrankadaptationlarge}, followed by Neftune \cite{jain2023neftunenoisyembeddingsimprove} for QoR enhancements. These optimized models are used in an HDL-Chat Model, which undergoes additional fine-tuning with specific instructions derived from the data augmentation steps. The fine-tuned model is evaluated against public benchmarks~\cite{thakur2023benchmarking} and \cite{liu2023verilogeval} within the evaluation pipeline. 

\section{Fine-Tuning Methodology}
This section outlined a series of fine-tuning experiments that we conducted to explore and enhance the performance of various code-generation LLMs. Our experiments involve the use of multiple fine-tuned candidate models, including StarCoder (SC), StarCoder2 (SC2), and CodeLlama (CL-FT) utilizing the PEFT technique. We replicated the CodeGen-16B model (CG) following~\cite{thakur2024verigen} and verified results versus published results in ~\cite{thakur2023benchmarking} for our comparisons. The QoR enhancement of each candidate model was then tested after performing fine-tuning using the Neftune technique. 
\begin{enumerate}
    \item \textbf{Experiment 1 - Base Model Selection:} The first experiment involved the application of PEFT LoRA fine-tuning to four distinct models: StarCoder, StarCoder2, Codegen, and CodeLLama. The initial version of augmented data was used for this experiment. 
    \item \textbf{Experiment 2 - Fine-Tuning with PEFT:} The second experiment centered around PEFT-based fine-tuning of SC and SC2 using the final version of augmented data. This data version embodied all data pipeline steps as depicted in Figure 1. The goal was to enhance the models' capability across four crucial tasks: code generation, code explanation, bug finding/fixing, and writing testbenches.
    \item \textbf{Experiment 3 - Fine-Tuning with Neftune:} The third experiment further fine-tuned the SC model using the final augmented dataset along with an additional technique, Neftune, to bolster the Quality of Results (QoR). Neftune was introduced to add more variability during the fine-tuning phase and to assess the model's capacity for handling intricate code generation tasks.
\end{enumerate}
The StarCoder variants, SC and SC2, demonstrated better performance than CodeGen and CodeLlama. Therefore, we chose the StarCoder variants as base LLMs for HDL-GPT in our work. We derived three fine-tuned models at the end of the model selection experiments, which are used in the benchmarking and analysis experiments described in Section~\ref{expt}: 
\begin{enumerate}
    \item \textbf{HDL-GPT (HG)} is a fine-tuned version of StarCoder.
    \item \textbf{HDL-GPT2 (HG2)} is a fine-tuned version of StarCoder2.
    \item \textbf{CL-FT} is a fine-tuned version of CodeLlama.
\end{enumerate}

\section{Experiments and Results}
\label{expt}
We evaluated HDL-GPT on various tasks including HDL code generation, explanation, bug finding, bug fixing, and test-bench (TB) generation. For the code generation task, we use NYU \cite{thakur2023benchmarking} and NVIDIA \cite{liu2023verilogeval} benchmarks and pass@k as the evaluation metric. For all other tasks, we use student-teacher grading~\cite{agarwal2024copilotevaluationharnessevaluating} as the quality metric with GPT-4 acting as the teacher. All experiments were conducted on NVIDIA A100 machines using  8 X 80GB GPUs. 

\subsection{Code Generation - NYU Benchmark}
Table \ref{table:comparison_NYU_eval_set_1} shows the comparison of four models using the NYU Eval Set 1 \cite{thakur2023benchmarking}. The NYU Eval Set 1 was used to compare four models: CG, HDL-GPT, HDL-GPT2, and CL-FT at three difficulty levels. HDL-GPT scored maximum points in Basic and Advanced levels with an average score of \textbf{0.92}. HDL-GPT2 outperformed others across all levels, achieving the highest average score of \textbf{0.96}. CodeLlama FT performed comparably at the Basic level and better at higher levels than CG, yielding an average score of 0.58.
\begin{table}[!htbp]  
\centering  
\caption{Comparison of Models based on NYU Eval Set 1}  
\label{table:comparison_NYU_eval_set_1}  
\vspace{-0.1cm}
\begin{tabular}{|p{1.7cm}|p{1.2cm}|p{1.2cm}|p{1.2cm}|p{1.2cm}|}
\hline  
\textbf{Difficulty} & \textbf{CG} & \textbf{HG} & \textbf{HG2} & \textbf{CL-FT} \\ \hline  
Basic & 0.745 & \textbf{1} & \textbf{1} & 0.75 \\ \hline  
Intermediate & 0.27 & \textbf{0.75} & \textbf{0.875} & 0.5 \\ \hline  
Advanced & 0.29 & \textbf{1} & \textbf{1} & 0.5 \\ \hline  
\textbf{Average} & 0.44 & \textbf{0.92} & \textbf{0.96} & 0.58 \\ \hline
\end{tabular}  
\end{table}  

The table \ref{table:comparison_NYU_eval_set_2} presents a comparison of four models: CG, HDL-GPT, HDL-GPT2, and CL-FT based on the NYU Eval Set II \cite{thakur2023benchmarking} across several categories. 
HDL-GPT performs strongly across many categories compared to CG and CL-FT, with an average score of \textbf{0.67} and HDL-GPT2 performs the best with an average score of \textbf{0.73}. CL-FT generally scores lower, resulting in an average of 0.36.
\begin{table}[!ht]  
\centering  
\caption{Comparison of Models based on NYU Eval Set II}  
\label{table:comparison_NYU_eval_set_2}  
\vspace{-0.2cm}
\begin{tabular}{|p{2.5cm}|p{1.0cm}|p{1.0cm}|p{1.0cm}|p{1.0cm}|}  
\hline  
\textbf{Problem} & \textbf{CG} & \textbf{HG} & \textbf{HG2} & \textbf{CL-FT} \\ \hline  
Getting Started       & 1    & \textbf{1}    & \textbf{1}    & 1    \\ \hline
Basics                & 1  & \textbf{1}    & \textbf{1}    & 0.63 \\ \hline
Vectors               & 0.44  & \textbf{0.78} & \textbf{0.89} & 0.33 \\ \hline
Module Hierarchy      & \textbf{0.38} & 0.33 & 0.33 & 0.11 \\ \hline
Procedures            & 0.75 & \textbf{0.88} & \textbf{0.88} & 0.5  \\ \hline
More Features         & 0.57  & \textbf{0.63} & \textbf{0.63} & 0.13 \\ \hline
Basic Gates           & 0.76 & \textbf{0.88} & \textbf{0.88} & 0.59 \\ \hline
Multiplexers          & 0.8  & \textbf{1}    & \textbf{1}    & 0.8  \\ \hline
Arithmetic Circuits   & 0.57  & \textbf{0.57} & \textbf{0.57} & 0.43 \\ \hline
K-Map to Circuits     & 0.13 & \textbf{0.75} & \textbf{0.75} & 0.63 \\ \hline
Latches \& Flip-flops & 0.5 & \textbf{0.83} & \textbf{0.94} & 0.33 \\ \hline
Counters              & 0.5 & \textbf{0.5}  & \textbf{0.5}  & 0.38 \\ \hline
Shift Registers       & 0.11    & \textbf{0.33} & \textbf{0.44} & 0    \\ \hline
Cellular Automata     & 0.33    & \textbf{0.33} & \textbf{0.67} & 0    \\ \hline
FSM                   & 0.24 & \textbf{0.48} & \textbf{0.61} & 0    \\ \hline
Larger Circuits       & 0.14    & \textbf{0.57} & \textbf{0.71} & 0.14 \\ \hline
Find bugs             & 0.4  & \textbf{0.6}  & \textbf{0.6}  & 0.2  \\ \hline
\textbf{Average}      & 0.51 & \textbf{0.67} & \textbf{0.73} & 0.36  \\ \hline
\end{tabular}  
\end{table}

\begin{table}[!ht]  
\centering  
\caption{NVIDIA Human Eval results (temperature=0.8)}  
\label{table:human_eval_combined}  
\vspace{-0.2cm}
\begin{tabular}{|c|c|c|c|}  
\hline  
\textbf{Model Versions} & \textbf{pass@1} & \textbf{pass@5} & \textbf{pass@10} \\ \hline  
GPT-3.5 & 26.7 & 45.8 & 51.7 \\ \hline  
GPT-4 & 27.0 & 45.8 & 52.0 \\ \hline  
verilog-sft & 28.8 & 45.9 & 52.3 \\ \hline  
HDL-GPT & \textbf{43.0} & \textbf{63.5} & \textbf{70.5} \\ \hline  
HDL-GPT2 & \textbf{60.6} & \textbf{78.5} & \textbf{81.4} \\ \hline  
\end{tabular}  
\end{table}
\subsection{Code Generation - NVIDIA Benchmark}
Table \ref{table:human_eval_combined} compares various models on NVIDIA Human Eval \cite{liu2023verilogeval} including OpenAI GPT-3.5, GPT-4, verilog-sft \cite{liu2023verilogeval}, HDL-GPT, and HDL-GPT2 at a temperature of 0.8. 

We have directly cited the verilog-sft model's metrics on these benchmarks as published in~\cite{liu2023verilogeval} for comparisons as we do not have direct access to the model. The remaining models are evaluated on three metrics: pass@1, pass@5, and pass@10. HDL-GPT significantly outperformed OpenAI GPT3.5, GPT4, and verilog-sft models at all pass rates. HDL-GPT2 delivered the highest performance surpassing HDP-GPT, with pass rates of \textbf{60.6\%} at pass@1, \textbf{78.5\%} at pass@5, and \textbf{81.4\%} at pass@10.
 
\begin{table}[!ht]  
\centering  
\caption{NVIDIA Machine Eval results (temperature=0.8)}  
\label{table:machine_eval}  
\vspace{-0.2cm}
\begin{tabular}{|c|c|c|c|}  
\hline  
\textbf{Model Versions} & \textbf{pass@1} & \textbf{pass@5} & \textbf{pass@10} \\ \hline  
GPT-3.5 & 46.7 & 69.1 & 74.1 \\ \hline  
GPT-4 & 47.9 & 67.8 & 72.9 \\ \hline  
verilog-sft & 46.2 & 67.3 & 73.7 \\ \hline  
HDL-GPT & \textbf{48.6} & \textbf{71.9} & \textbf{77.6} \\ \hline  
HDL-GPT2 & \textbf{67.0} & \textbf{86.0} & \textbf{90.2} \\ \hline  
\end{tabular}  
\vspace{-0.45cm}
\end{table} 

Table \ref{table:machine_eval} compares all the models at a temperature of 0.8 on the machine evaluation benchmark.
HDL-GPT showed better performance with \textbf{48.6\%} for pass@1, \textbf{71.9\%} for pass@5, and \textbf{77.6\%} for pass@10. Similarly, HDL-GPT2 shows even better performance and outperformed all with pass rates of \textbf{67.0\%} for pass@1, \textbf{86.0\%} for pass@5, and \textbf{90.2\%} for pass@10. 

\subsection{Code Analysis tasks}
\begin{figure}[h!tbp]
	\centering
	\includegraphics[width=1\columnwidth]{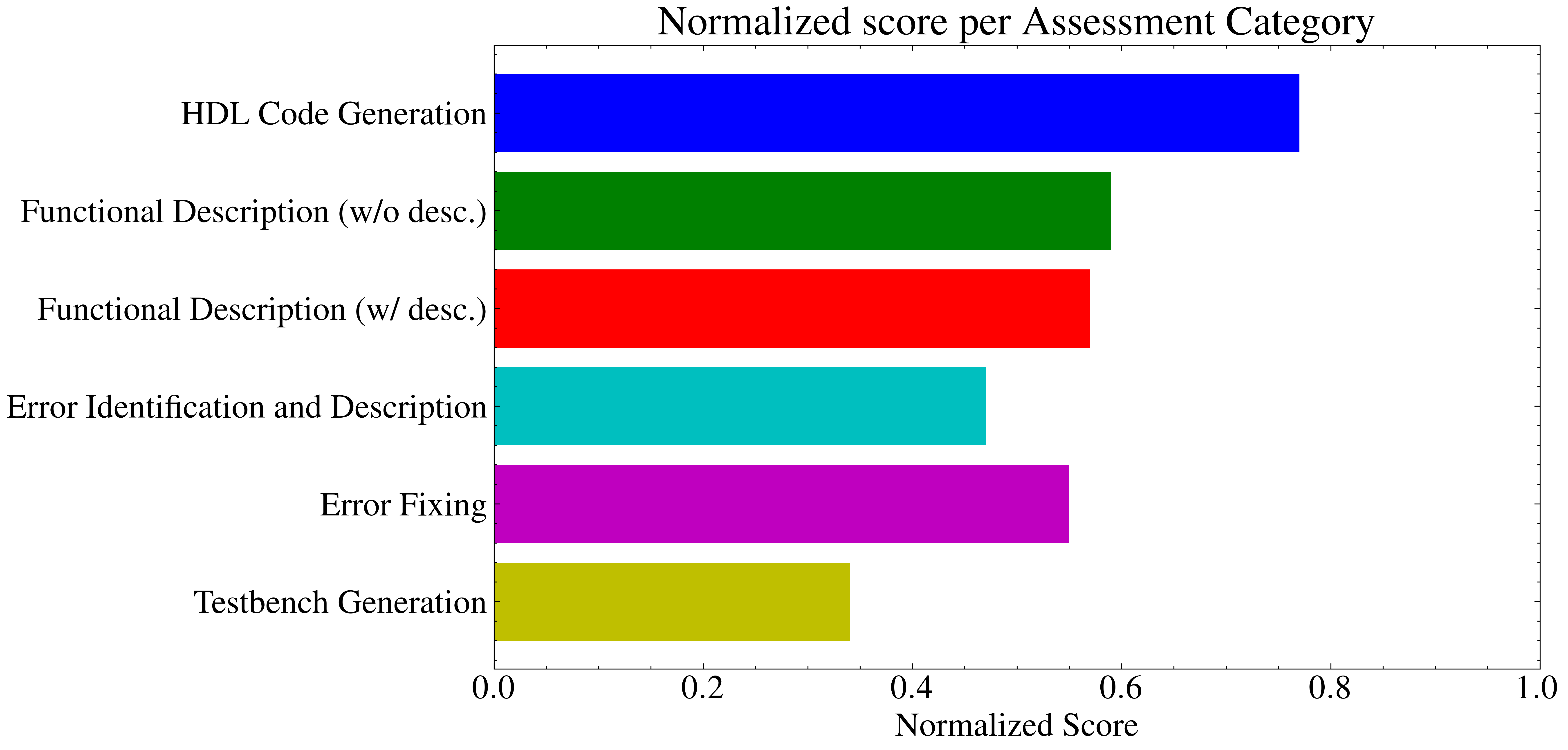}
    \vspace{-0.45cm}
	\caption{HDL-GPT Normalized Student-Teacher Scores}
	\label{fig_6}
\end{figure}

We use a student-teacher grading evaluation scheme on the same benchmarks as in the prior experiments for code analysis, bug finding, bug fixing, and testbench generation tasks. We evaluated four LLMs: Mixtral, Llama, OpenAI GPT-3.5, and GPT-4 as teacher candidates. The teacher model is prompted to perform scoring on a scale of 0-5 on the given HDL code based on detailed description, code quality, and code correctness. We then calculate the average grading scores (0-5) for each evaluation set and normalize to a range between 0 and 1 as shown in Equation \ref{eq1}. This is the normalized score reported in experiments in Figure \ref{fig_6}.
\begin{equation}
    \text{{Normalized Score}} = \frac{{\text{{Average Score for a Eval Set}} }}{{\text{{Maximum Score}} }}
    \label{eq1}
\end{equation}

We selected the best teacher model (GPT-4) by hand-curating grades across all teacher candidates on a random sample of 150 tasks from the evaluation dataset.
Figure \ref{fig_6} shows the grading results for the Student-Teacher grading in various categories. The HDL code generation task received a normalized score of \textbf{0.77}. The functional description task of undocumented HDL code had a normalized score of \textbf{0.59} and \textbf{0.57} with the actual description. The task of identifying and describing errors received a normalized score of \textbf{0.47} while the Error fixing task achieved a normalized score of \textbf{0.55}. The test-bench generation task had the lowest normalized score of 0.34 showing room for improvement for HDL-GPT.

\section{Conclusion}
The work in this paper underscores the critical role of high-quality training data, input prompts, and fine-tuning methods (PEFT and Neftune) in creating a SOTA LLM, called HDL-GPT, for HDL code generation, explanation, bug detection/repair, and testbench creation. The experimental results against public benchmarks demonstrate that the model's performance, as well as the quality of the generated code, rely heavily on the training data's quality. Our future research will delve into optimizing HDL-GPT's performance for IC design flows, creating representative benchmark suites for IC design tasks, and enhancing its potential for high-quality code generation and task performance in the IC design flows.

\bibliographystyle{./bibliography/IEEEtran}
\bibliography{./bibliography/IEEEabrv,./bibliography/references}

\begin{thebibliography}{10}
\providecommand{\url}[1]{#1}
\csname url@samestyle\endcsname
\providecommand{\newblock}{\relax}
\providecommand{\bibinfo}[2]{#2}
\providecommand{\BIBentrySTDinterwordspacing}{\spaceskip=0pt\relax}
\providecommand{\BIBentryALTinterwordstretchfactor}{4}
\providecommand{\BIBentryALTinterwordspacing}{\spaceskip=\fontdimen2\font plus
\BIBentryALTinterwordstretchfactor\fontdimen3\font minus \fontdimen4\font\relax}
\providecommand{\BIBforeignlanguage}[2]{{%
\expandafter\ifx\csname l@#1\endcsname\relax
\typeout{** WARNING: IEEEtran.bst: No hyphenation pattern has been}%
\typeout{** loaded for the language `#1'. Using the pattern for}%
\typeout{** the default language instead.}%
\else
\language=\csname l@#1\endcsname
\fi
#2}}
\providecommand{\BIBdecl}{\relax}
\BIBdecl

\bibitem{mack2015multiple}
C.~Mack, ``The multiple lives of moore's law,'' \emph{IEEE Spectrum}, vol.~52, no.~4, pp. 31--31, 2015.

\bibitem{openai2024gpt4technicalreport}
\BIBentryALTinterwordspacing
OpenAI, J.~Achiam, S.~Adler, S.~Agarwal, L.~Ahmad, I.~Akkaya, F.~L. Aleman, D.~Almeida, J.~Altenschmidt, S.~Altman, S.~Anadkat \emph{et~al.}, ``{GPT-4} technical report,'' 2024. [Online]. Available: \url{https://arxiv.org/abs/2303.08774}
\BIBentrySTDinterwordspacing

\bibitem{jiang2024mixtralexperts}
\BIBentryALTinterwordspacing
A.~Q. Jiang, A.~Sablayrolles, A.~Roux, A.~Mensch, B.~Savary, C.~Bamford, D.~S. Chaplot, D.~de~las Casas, E.~B. Hanna, F.~Bressand, G.~Lengyel, G.~Bour, G.~Lample, L.~R. Lavaud, L.~Saulnier, M.-A. Lachaux, P.~Stock, S.~Subramanian, S.~Yang, S.~Antoniak, T.~L. Scao, T.~Gervet, T.~Lavril, T.~Wang, T.~Lacroix, and W.~E. Sayed, ``Mixtral of experts,'' 2024. [Online]. Available: \url{https://arxiv.org/abs/2401.04088}
\BIBentrySTDinterwordspacing

\bibitem{touvron2023llama2openfoundation}
\BIBentryALTinterwordspacing
H.~Touvron, L.~Martin, K.~Stone, P.~Albert, A.~Almahairi, Y.~Babaei, N.~Bashlykov, S.~Batra, P.~Bhargava, S.~Bhosale \emph{et~al.}, ``Llama 2: Open foundation and fine-tuned chat models,'' 2023. [Online]. Available: \url{https://arxiv.org/abs/2307.09288}
\BIBentrySTDinterwordspacing

\bibitem{tsai2024rtlfixerautomaticallyfixingrtl}
\BIBentryALTinterwordspacing
Y.-D. Tsai, M.~Liu, and H.~Ren, ``{RTLFixer}: Automatically fixing {RTL} syntax errors with large language models,'' 2024. [Online]. Available: \url{https://arxiv.org/abs/2311.16543}
\BIBentrySTDinterwordspacing

\bibitem{gunasekar2023textbooksneed}
\BIBentryALTinterwordspacing
S.~Gunasekar, Y.~Zhang, J.~Aneja, C.~C.~T. Mendes, A.~D. Giorno, S.~Gopi, M.~Javaheripi, P.~Kauffmann, G.~de~Rosa, O.~Saarikivi, A.~Salim \emph{et~al.}, ``Textbooks are all you need,'' 2023. [Online]. Available: \url{https://arxiv.org/abs/2306.11644}
\BIBentrySTDinterwordspacing

\bibitem{thakur2024verigen}
S.~Thakur, B.~Ahmad, H.~Pearce, B.~Tan, B.~Dolan-Gavitt, R.~Karri, and S.~Garg, ``Verigen: A large language model for verilog code generation,'' \emph{ACM Transactions on Design Automation of Electronic Systems}, vol.~29, no.~3, pp. 1--31, 2024.

\bibitem{liu2024chipnemodomainadaptedllmschip}
\BIBentryALTinterwordspacing
M.~Liu, T.-D. Ene, R.~Kirby, C.~Cheng, N.~Pinckney, R.~Liang, J.~Alben, H.~Anand, S.~Banerjee, I.~Bayraktaroglu, B.~Bhaskaran \emph{et~al.}, ``{ChipNeMo: Domain-Adapted LLMs for Chip Design},'' 2024. [Online]. Available: \url{https://arxiv.org/abs/2311.00176}
\BIBentrySTDinterwordspacing

\bibitem{li2023starcodersourceyou}
\BIBentryALTinterwordspacing
R.~Li, L.~B. Allal, Y.~Zi, N.~Muennighoff, D.~Kocetkov, C.~Mou, M.~Marone, C.~Akiki, J.~Li, J.~Chim, Q.~Liu \emph{et~al.}, ``{StarCoder}: may the source be with you!'' 2023. [Online]. Available: \url{https://arxiv.org/abs/2305.06161}
\BIBentrySTDinterwordspacing

\bibitem{hoffmann2022trainingcomputeoptimallargelanguage}
\BIBentryALTinterwordspacing
J.~Hoffmann, S.~Borgeaud, A.~Mensch, E.~Buchatskaya, T.~Cai, E.~Rutherford, D.~de~Las~Casas, L.~A. Hendricks, J.~Welbl, A.~Clark, T.~Hennigan \emph{et~al.}, ``Training compute-optimal large language models,'' 2022. [Online]. Available: \url{https://arxiv.org/abs/2203.15556}
\BIBentrySTDinterwordspacing

\bibitem{wei2022chain}
J.~Wei, X.~Wang, D.~Schuurmans, M.~Bosma, F.~Xia, E.~Chi, Q.~V. Le, D.~Zhou \emph{et~al.}, ``Chain-of-thought prompting elicits reasoning in large language models,'' \emph{Advances in neural information processing systems}, vol.~35, pp. 24\,824--24\,837, 2022.

\bibitem{roziere2024codellamaopenfoundation}
\BIBentryALTinterwordspacing
B.~Rozière, J.~Gehring, F.~Gloeckle, S.~Sootla, I.~Gat, X.~E. Tan, Y.~Adi, J.~Liu, R.~Sauvestre, T.~Remez, J.~Rapin \emph{et~al.}, ``{Code Llama: Open Foundation Models for Code},'' 2024. [Online]. Available: \url{https://arxiv.org/abs/2308.12950}
\BIBentrySTDinterwordspacing

\bibitem{hu2021loralowrankadaptationlarge}
\BIBentryALTinterwordspacing
E.~J. Hu, Y.~Shen, P.~Wallis, Z.~Allen-Zhu, Y.~Li, S.~Wang, L.~Wang, and W.~Chen, ``{LoRA: Low-Rank Adaptation of Large Language Models},'' 2021. [Online]. Available: \url{https://arxiv.org/abs/2106.09685}
\BIBentrySTDinterwordspacing

\bibitem{jain2023neftunenoisyembeddingsimprove}
\BIBentryALTinterwordspacing
N.~Jain, P.~yeh Chiang, Y.~Wen, J.~Kirchenbauer, H.-M. Chu, G.~Somepalli, B.~R. Bartoldson, B.~Kailkhura, A.~Schwarzschild, A.~Saha, M.~Goldblum, J.~Geiping, and T.~Goldstein, ``Neftune: Noisy embeddings improve instruction finetuning,'' 2023. [Online]. Available: \url{https://arxiv.org/abs/2310.05914}
\BIBentrySTDinterwordspacing

\bibitem{thakur2023benchmarking}
S.~Thakur, B.~Ahmad, Z.~Fan, H.~Pearce, B.~Tan, R.~Karri, B.~Dolan-Gavitt, and S.~Garg, ``Benchmarking large language models for automated verilog {RTL} code generation,'' in \emph{2023 Design, Automation \& Test in Europe Conference \& Exhibition (DATE)}.\hskip 1em plus 0.5em minus 0.4em\relax IEEE, 2023, pp. 1--6.

\bibitem{liu2023verilogeval}
M.~Liu, N.~Pinckney, B.~Khailany, and H.~Ren, ``{VerilogEval: Evaluating Large Language Models for Verilog Code Generation},'' in \emph{2023 IEEE/ACM International Conference on Computer Aided Design (ICCAD)}.\hskip 1em plus 0.5em minus 0.4em\relax IEEE, 2023, pp. 1--8.

\bibitem{agarwal2024copilotevaluationharnessevaluating}
\BIBentryALTinterwordspacing
A.~Agarwal, A.~Chan, S.~Chandel, J.~Jang, S.~Miller, R.~Z. Moghaddam, Y.~Mohylevskyy, N.~Sundaresan, and M.~Tufano, ``Copilot evaluation harness: Evaluating llm-guided software programming,'' 2024. [Online]. Available: \url{https://arxiv.org/abs/2402.14261}
\BIBentrySTDinterwordspacing

\end{thebibliography}


\end{document}